# A XGBOOST RISK MODEL VIA FEATURE SELECTION AND BAYESIAN HYPER-PARAMETER OPTIMIZATION

Yan Wang[1], Xuelei Sherry Ni[2]

[1]Graduate College, Kennesaw State University, Kennesaw, USA
ywang63@students.kennesaw.edu
[2]Department of Statistics and Analytical Sciences, Kennesaw State University, Kennesaw, USA
sni@kennesaw.edu

## ABSTRACT

*This paper aims to explore models based on the extreme gradient boosting (XGBoost) approach for business risk classification. Feature selection (FS) algorithms and hyper-parameter optimizations are simultaneously considered during model training. The five most commonly used FS methods including weight by Gini, weight by Chi-square, hierarchical variable clustering, weight by correlation, and weight by information are applied to alleviate the effect of redundant features. Two hyper-parameter optimization approaches, random search (RS) and Bayesian tree-structured Parzen Estimator (TPE), are applied in XGBoost. The effect of different FS and hyper-parameter optimization methods on the model performance are investigated by the Wilcoxon Signed Rank Test. The performance of XGBoost is compared to the traditionally utilized logistic regression (LR) model in terms of classification accuracy, area under the curve (AUC), recall, and F1 score obtained from the 10-fold cross validation. Results show that hierarchical clustering is the optimal FS method for LR while weight by Chi-square achieves the best performance in XG-Boost. Both TPE and RS optimization in XGBoost outperform LR significantly. TPE optimization shows a superiority over RS since it results in a significantly higher accuracy and a marginally higher AUC, recall and F1 score. Furthermore, XGBoost with TPE tuning shows a lower variability than the RS method. Finally, the ranking of feature importance based on XGBoost enhances the model interpretation. Therefore, XGBoost with Bayesian TPE hyper-parameter optimization serves as an operative while powerful approach for business risk modeling.*

## KEYWORDS

*Extreme gradient boosting; XGBoost; feature selection; Bayesian tree-structured Parzen estimator; risk modeling*

## 1. INTRODUCTION

Risk modeling is an effective tool to assist financial institutions to properly decide whether or not to grant loans to business or other applicants [1]. Thereby, the problem of risk modeling is transformed into a binary classification task, i.e., grant loans to low risk applicants or not grant to those with high risk. Logistic regression (LR) is a traditionally utilized technique for binary classifications in the financial domain because of its easy implementation, explainable results, as well as the similar and often better performance compared to other binary classifiers such as decision trees and neural networks [2] [3] [4] [5] [6]. On the other hand, it has been shown that a single classifier cannot solve all problems effectively while ensemble models have been revealed to be promising in many credit risk studies [7] [8] [9]. One of the state-of-the-art ensemble approach is the extreme gradient boosting (XGBoost). It is a novel while advanced variant of the gradient boosting algorithm and has obtained promising results in many Kaggle machine learning competitions [10]. Furthermore, XGBoost has been successfully applied in bankruptcy prediction and credit scoring in a few studies [11][12].

Numerous studies have focused on offering novel mechanisms to enhance the performance of credit risk modeling. It has been demonstrated that feature selection (FS) is one of the efficient approaches in improving model performance because of its ability to alleviate the effects of noise and redundant variables [13]. Another method for model-improving is the hyper-parameter optimization or tuning. It is shown that careful hyper-parameter tuning tends to prevent the failure and reduce the over-fitting problem of XGBoost. The two main strategies used for finding the proper setting of hyper-parameters in XGBoost are random search (RS) and Bayesian tree-structured Parzen estimator (TPE). They have demonstrated substantial influence on classification performance [14] [15].

After careful paper review, we find that there is seldom research aiming at exploring the effect of FS and hyper-parameter optimizations simultaneously on XGBoost in the financial domain. Therefore, motivated by the aforementioned studies, we set up a series of experiments that contain FS methods and hyper-parameter optimizations simultaneously, thereby exploring an accurate and comprehensive business risk model based on XGBoost. The superiority of XGBoost over the widely used LR is evaluated via classification accuracy, area under the curve (AUC), recall, and F1 score. Moreover, the effect of different FS as well as hyper-parameter optimization methods on the model performance is comprehensively investigated through the Wilcoxon signed rank test. Finally, the features are ranked according to their importance score to enhance the model interpretation.

This paper has been structured as follows. Since different FS methods and XGBoost models along with the hyper-parameter optimization are used in this study, we will first describe the relevant algorithms in Section 2. Then the experimental design is discussed in Section 3. Section 4 demonstrates the experimental results and discussions. Finally, Section 5 addresses the conclusions.

## 2. ALGORITHMS

In this section, the algorithms related to FS and XGBoost along with hyper-parameter optimizations are discussed.

### 2.1. Feature selection methods

FS methods aims to filter the redundant variables and select the most appropriate subset of features. By applying FS methods to the dataset, we can decrease the effect of the noise as well as reduce the computational cost during the modeling stage. Many studies have shown that FS can be used to increase the classification performance [13] [16].

In this study, five commonly used FS methods are applied and evaluated: weight by gini index, weight by chi-square, hierarchical variable clustering, weight by correlation, and weight by information gain ratio. For simplicity, we use the terms with initial capitalization to denote different FS methods. Therefore, Gini, Chi-square, Cluster, Correlation, and Information are used to represent the aforementioned five FS approaches, respectively. In the Gini FS method, the value of an attribute is evaluated via the gini impurity index. Similarly, Chi-square, Correlation, and Information evaluates the relevance of the feature by calculating its chi-squared statistic, correlation, and information gain ratio with respect to the target variable [17]. Features with higher values of gini index, chi-squared statistic, correlation, and information gain ratio are selected in the FS results. On the other hand, the Cluster method bases on the variable clustering analysis and selects the best feature within each cluster according to the 1-$R^2$ ratio defined in Eq. 1 [18]. Different from the rest of the four FS methods, features with lower 1-$R^2$ ratio are selected by the Cluster method.

$$1 - R^2 \text{ ratio } = \frac{1 - R^2_{own\_cluster}}{1 - R^2_{next\_closest\_cluster}} \tag{1}$$

## 2.2. Logistic Regression

LR is a standard binary classification technique widely used in industry because of its simplicity and balanced error distribution [19] [21]. It outputs the conditional probability p of an observation that belongs to a specific class using the formula defined in Eq. 2, where ($x_1, x_2, ..., x_k$) denotes the input variables while ($\beta_0..., \beta_k$) represents the unknown parameters that need to be estimated.

$$p = \left( \frac{\exp(\beta_0 + \beta_1 * x_1 + \beta_2 * x_2 + \cdots + \beta_k * x_k)}{1 + \exp(\beta_0 + \beta_1 * x_1 + \beta_2 * x_2 + \cdots + \beta_k * x_k)} \right) \tag{2}$$

## 2.3. Extreme gradient boosting along with hyper-parameters

XGBoost was proposed in 2015 and has been frequently applied because of its rapidness, efficiency, and scalability [10]. It is an advanced implementation of the gradient boosting (GB) algorithm and uses the decision tree as the base classifier. After carefully reading the research in [12] and [20], the algorithm of GB and XGBoost is briefly summarized as follows. Suppose we have a dataset D = {$x; y$} containing n observations, where $x$ and $y$ denotes the features and the target variable, respectively. In GB, suppose there are K number of boosting, then we use B additive functions to predict the output. Denote $\hat{y}_i$ as the prediction for the $i$-th instance at the $b$-th boost, $f_b$ represents a tree structure q with leaf j having a weight score $w_j$. Then for a given instance $x_i$, the final prediction is calculated by summing up the scores across all leaves and this can be expressed in Eq. 3.

$$\hat{y}_i = \sum_{b=1}^{B} f_b(x_i) \tag{3}$$

The idea of GB is to minimize the loss function $L_b$ defined using Eq. 4, where l($y_i, \hat{y}_i$) measures the difference between the prediction and its real value $y_i$. Since the base learner of GB is decision tree, several hyper-parameters related to the tree structures including subsample, max leaves, and max depth are employed to reduce the over-fitting problem as well as to enhance the model. Moreover, learning rate or the shrinkage factor, which controls the weighting of new trees added to the model, is also used to decrease the rate of the model's adaptation to the training data. The above-mentioned hyper-parameters are also defined in XGBoost and their descriptions can be found in Table 1.

$$L_b = \sum_{i=1}^{n} l(y_i, \hat{y}_i) \tag{4}$$

By adding a regularization term $\Omega(f_b)$ to the loss function defined in Eq. 4, we can get the loss function of XGB described in Eq. 5, The regularization term $\Omega(f_b)$ penalizes the model complexity. It can be expressed by summing up two parts: $\gamma T$ and $0.5\,\lambda||w||^2$. $T$ represents the number of leaves that are contained by the tree. The hyper-parameter $\gamma$ defines the minimum loss reduction for further partition. If the loss reduction is less than $\gamma$, XGBoost stops, implying that penalizes the model complexity. $\lambda$ is a fixed coefficient and $||w||^2$ represents the L2 norm of the weight of the leaf. Similar to $\gamma$, a hyper-parameter $w_{mc}$ controls the tree depth and a substantial $w_{mc}$ makes the model more conservative in splitting. $w_{mc}$ is defined as the minimum sum of the in-stance weight in further partitioning. The descriptions of the above-mentioned hyper-parameters can be found in Table 1.

$$L_b = \sum_{i=1}^{n} l(y_i, \hat{y}_i) + \sum_{b=1}^{B} \Omega(f_b) = \sum_{i=1}^{n} l(y_i, \hat{y}_i) + \gamma T + 0.5\,\lambda||w||^2 \quad (5)$$

Compared with GB, another technique used in XGBoost for the further prevention of over-fitting is the column subsampling or feature subsampling [11]. It is shown that using column subsampling is even more efficient than traditional row subsampling in preventing over-fitting [14]. The description of the corresponding hyper-parameter "colsample_bytree" can be found in Table 1.

### 2.4. Hyper-parameter optimization methods in XGBoost

In XGB, hyper-parameter optimization (i.e., tuning) aims at searching for the hyper-parameter values that minimizes the objective function defined in Eq. 5. There are two popular hyper-parameter optimization methods: RS and Bayesian Tree Parzen Estimators (TPE). RS means the hyper-parameters are randomly picked from the pre-defined searching domain uniformly and the searching does not depend on the previous boosting result [14] [22]. It has been shown to be efficient for problems with high dimensions in some studies. On the contrary, Sequential Model Based Optimization (SMBO), which is also named Bayesian optimization, is a probability based approach and uses the probability model to select the most promising hyper-parameters [23]. According to the choices of the probability model (i.e., the surrogate model), several variants of SMBO are proposed including Gaussian Processes, Random Forest Regressions, and TPE [24] [25]. Since several studies have revealed the promising results via TPE approach, we adopt this method in our study [26] [27]. For simplicity, in the rest of the paper, we use XGB_TPE and XGB_RS to denote the XGBoost models built by using Bayesian TPE and RS hyper-parameter optimization methods, respectively.

## 3. EXPERIMENTAL DESIGN

In this study, we aim to answer the following four research questions explicitly based on the dataset used:

- How different FS methods affect the performance of LR and XGBoost? What is the corresponding optimal FS method for different models?

- How the hyper-parameter optimization methods including RS and TPE affect the performance of XGBoost?

- Is the XGBoost method more powerful in business risk prediction compared to traditionally utilized LR?

- Based on the dataset used in this study, what are the important features in the risk prediction?

To address the above-mentioned questions, a comprehensive experimental study is conducted and the details are described in the following subsections.

### 3.1. Data Description

The dataset used in this study is contributed by a national credit bureau and contains over 10 million de-identified commercial information of the companies in the U.S. from 2006 to 2014. The 305 independent variables are all numeric and provide information of the companies' activities in non-financial accounts, telecommunication accounts, and industry accounts, etc. The dependent variable *RiskFlag* represents whether the business is in risk or not. The positive rate (i.e., proportion of risky business) in the dataset is about 52%.

### 3.2. Data Pre-processing

Based on the original dataset, we first replaced the invalid records of variables with missing values and then removed the variables with missing percentage larger than 70%. As the result, out of the 305 independent variables, 108 variables are kept in our study. Then a stratified sampling procedure was applied to obtain a sample with 8000 observations for the further experiments described in this Section.

### 3.3. Searching domain of hyper-parameters in XGBoost

As discussed in Sections 2.3 and 2.4, several hyper-parameters of XGBoost needs to be optimized based on a pre-defined domain using RS and Bayesian TPE methods to avoid the over-fitting in this study. Although many hyper-parameters are included in XGBoost, we only focus on those that are shown to have significant effect on the model performance in the previous studies. The hyper-parameters adopted in this study include "learning rate", "subsample", "max_leaves", "max_depth", "gamma", "colsample_bytree", and "min_child_weight". The corresponding searching domain and the descriptions of the hyper-parameters are summarized in Table 1. The settings of the searching domain are based on the suggestions from previous research as well as based on our initial trials [28] [29] [30]. For the rest of other hyper-parameters including "n_estimators" (number of boosted trees), "min_child_samples" (minimum number of data needed in a leaf) and "subsample_for_bin" (number of samples for constructing bins), we use the default settings in Python [20].

### 3.4. Performance evaluation criteria

The criteria used to evaluate the model performance are discussed in this section. Accuracy is the commonly used measure in binary classification problems and can provide reasonable model comparisons [31]. In this study, True Positive (TP) and False Positive (FP) represent correctly and wrongly classified risky businesses, respectively. True Negative (TN) and False Negative (FN) denote correctly and wrongly classified non-risky businesses, respectively. Then accuracy can be calculated using Eq. 6. Another evaluation measure used, AUC, is the area under the Receiver Operating Characteristic Curve (ROC) since it measures how well the model distinguishes the positives and the negatives [32]. ROC is plotted by using false positive rate (i.e., $\frac{FP}{FP+TN}$) on the x-axis and true positive rate (i.e., $\frac{TP}{TP+FN}$) on the y-axis. Recall (i.e., true positive rate) measures the fraction of positives that have been retrieved over the total amount of all the positives (defined in Eq. 7) while precision denotes the fraction of positives among the retrieved positives [33] (defined in Eq. 8). As discussed in [34], recall and precision are emphasized differently in risk modeling and hazard research domain. Similarly, in our study, recall is weighted more heavily than precision since a false negative error may signify the loss. F1 score (defined in Eq. 9) is another model evaluation measure in this paper since it is the harmonic mean of precision and recall [35].

Table 1. Searching domain of hyper-parameters in XGBoost. Hyper-parameters in bold are those that are defined only in XGBoost but not in GB.

| Name | Description | Domain |
| --- | --- | --- |
| learning rate | Step size shrinkage used in model update | (0.005, 0.2) |
| subsample | Subsample ratio of the training instances used for fitting the individual tree | (0.8, 1) |
| max_leaves | Maximum number of nodes to be added | (10, 200) |
| max_depth | Maximum depth of a tree | (5, 30) |
| **gamma ($\gamma$)** | Minimum loss reduction required for further partition | (0, 0.02) |
| **colsample_bytree** | Subsample ratio of features/columns used for fitting the individual tree | (0.8, 1) |
| **min_child_weight ($w_{mc}$)** | Minimum weights of the instances required in a leaf | (0, 10) |

$$\text{accuracy} = \frac{TP + TN}{TP + TN + FP + FN} \quad (6)$$

$$\text{recall} = \frac{TP}{TP + FN} \quad (7)$$

$$\text{precision} = \frac{TP}{TP + FP} \quad (8)$$

$$\text{F1 score} = \frac{2 * precision * recall}{precision + recall} \quad (9)$$

### 3.5. Flowchart of the experiments

After the data pre-processing procedures described in Section 3.2, the models are built on the training set while the performance is evaluated on the validation set. To ensure the reliability and accuracy of the results, we use 10-fold cross validation in this study. Fig.1 shows the flowchart of the analysis where a certain fold of the data is used as the validation set while the rest of the nine folds are used as the training set. As illustrated in Fig. 1, the entire analysis process contains six stages. In stage 1, the training set is pre-processed following the steps below:

- For each feature in the training set, we performed missing value imputation using its median value;
- Normalization of the variables by transforming every variable to its *z*-score using its mean and standard deviations in the training set.

In stage 2, five FS approaches including Gini, Chi-square, Cluster, Correlation, and Information are applied on the training set. This can select the most representative subset of the features. To make the comparison of the model performance based on different FS methods fair, we fix the size of the subset of the features as 50. The reason why we select 50 features is illustrated in Section 4.1. In stage 3, three models including LR, XGB_RS, and XGB_TPE are built using the subset of the features produced by different FS methods from stage 2. In stage 4, the validation set is pre-processed using similar strategies as that on the training set in stage 1. It is worth noting that for the pre-processing on the validation set, the median, mean, and standard deviation values should all come from the training set for each variable. Then in stage 5, the observations in the validation set are scored using the models obtained from stage 4. Finally, the model performance is evaluated using accuracy, AUC, recall, and F1 score in stage 6.

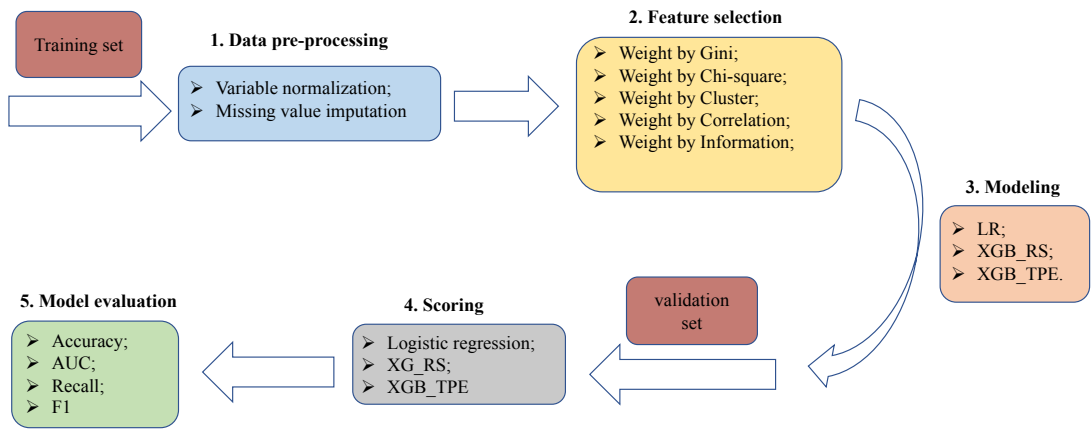

Figure 1. The flowchart of the experiments

## 3.6. Statistical significance analysis

To perform reasonable and reliable comparisons of different FS methods as well as model performance, we implement 10-fold cross validation 10 times in this study. For each of the four evaluation measures, the average value can be obtained from each of the 10-fold cross validation. Then after the implementation of the 10-time 10-fold cross validation, we can get 10 average values of the evaluation measure for each model based on a certain FS method. Take LR based on Gini using the accuracy measure as an example. By following the flowchart in Fig. 1, we get one accuracy value when a certain fold of the data is used as the validation set. After completing 10-fold cross validation for the first time, 10 accuracy values can be obtained. The average of the above-mentioned 10 accuracy values is recorded as the first average cross-validated accuracy. For the naming convention in this study, we record the evaluation results from the 10-fold cross validation using the format as follows: "FS_model_evaluation_index". Therefore, the first average cross-validated accuracy is denoted as Gini_LR_accuracy_1 in our analysis. After applying the 10-fold cross validation for 10 times, we get a series of values denoted as Gini_LR_accuracy_1, Gini_LR_accuracy_2,..., and Gini_LR_accuracy_10. Finally, the performance of LR based on Gini using the accuracy measure can be expressed by taking the average of the above-mentioned series of 10 values and is denoted as Gini_LR_accuracy. Furthermore, the stability and consistency of the model performance can be explored by calculating the sample standard deviation of these 10 values, which and is denoted by Gini_LR_accuracy_SD.

Pairwise Wilcoxon signed rank test, which is a non-parametric approach, is then employed to test the statistical significance of the differences in performance resulting by different methods. For example, by performing the paired Wilcoxon signed rank test on two series of values including (Gini_LR_accuracy_1, Gini_LR_accuracy_2,..., and Gini_LR_accuracy_10) and (Cluster_LR_accuracy_1, Cluster_LR_accuracy_2,..., and Cluster_LR_accuracy_10), we can examine whether Gini and Cluster can result in different accuracy in LR. By comparing the difference of the FS methods for each of the three models, the optimal FS approach for each model can be identified. Then, the pairwise Wilcoxon signed rank test is used to compare different evaluation criteria of the models comprehensively along with their optimal FS approach, thereby to select the final optimal model. With respect to the desired significant level during the pairwise comparison, it is set to $\alpha = 0.1$ in this study. Bonferroni correction is used in this study to handle the problem from the increased Type I error by testing each individual hypothesis [36]. As a result, each individual hypothesis is tested at the level $\alpha/m$, where m denotes the number of null hypothesis that are tested. For example, when comparing the performance of LR, XGB_RS, and XGB_TPE, three individual tests are needed and Bonferroni correction would test each individual hypothesis at $\alpha/3 = 0.033$.

## 4. RESULTS

In this section, the effects of different FS methods on model performance are demonstrated. Furthermore, Bayesian TPE hyper-parameter optimization on XGBoost is compared with the RS method. With respect to the analysis tools in this study, SAS (version 9.4) is used for data pre-processing that is labelled as stage 1 and 4 in Fig. 1. The Cluster FS method is also implemented in SAS and the rest four FS methods are implemented on RapidMiner (version 9.0). The training and scoring procedures of LR are implemented on RapidMiner as well. XGB_TPE and XGB_RS are performed on Python (version 3.5). All the experiments are operated on the desktop computer with MacOS system, 3.3 GHz Intel Core I7 process, and 16GB RAM.

### 4.1. Parameter settings in FS methods

In RapidMiner, one important parameter of FS that needs careful setting is "number of features selected". It is because too many features tend to hurt model performance due to the potential multicollinearity problem while too few features may not capture enough information based on the original dataset. In our study, the "number of features selected" is determined based on the result from the Cluster FS method. As shown in Fig. 2, over 90% of the variations in the original dataset can be explained by 50 clusters. Therefore, in the Cluster FS method, we select one representative feature from each of the 50 clusters and believe that enough information provided by the data can be kept. To ensure the fair comparison among different FS methods, the value of "number of features selected" is set to 50 for Gini, Chi-square, Correlation, and Information as well.

### 4.2. Best FS method in XGB_TPE

Fig. 3 demonstrated the XGB_TPE performance over five FS approaches by using the four evaluation measures. As described in Section 3.6, the experiments were implemented using 10-fold cross validation and were repeated 10 times, the evaluation measures expressed in Fig. 3 are the average cross-validated values along with the standard deviations. It is observed that different FS approaches produce very different results. The Chi-square method can achieve the highest accuracy, AUC, Recall and F1 score among the five FS methods. On the other hand, Gini has the worst performance since it results in the lowest values in any of the four evaluation criteria. There seems to be no obvious difference in the model performance between Chi-square and Cluster. The above-mentioned two FS methods outperform the rest three methods in all the evaluation measures. Moreover, the three FS methods including Gini, Correlation, and Information do not result in obvious difference in the model performance. Another finding is that, the small values of the standard deviations show the consistency and stability of the FS methods on the XGB_TPE model.

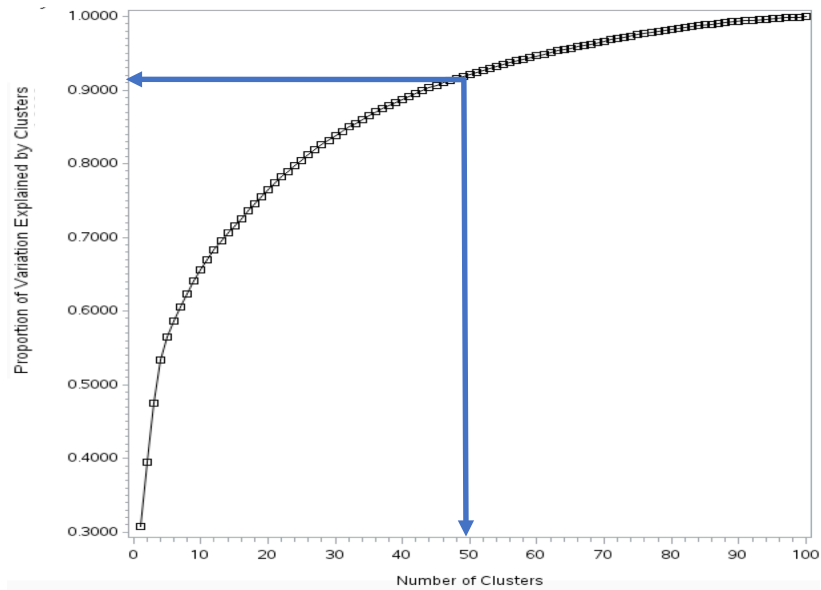

Figure 2. Plot of hierarchical variable clustering

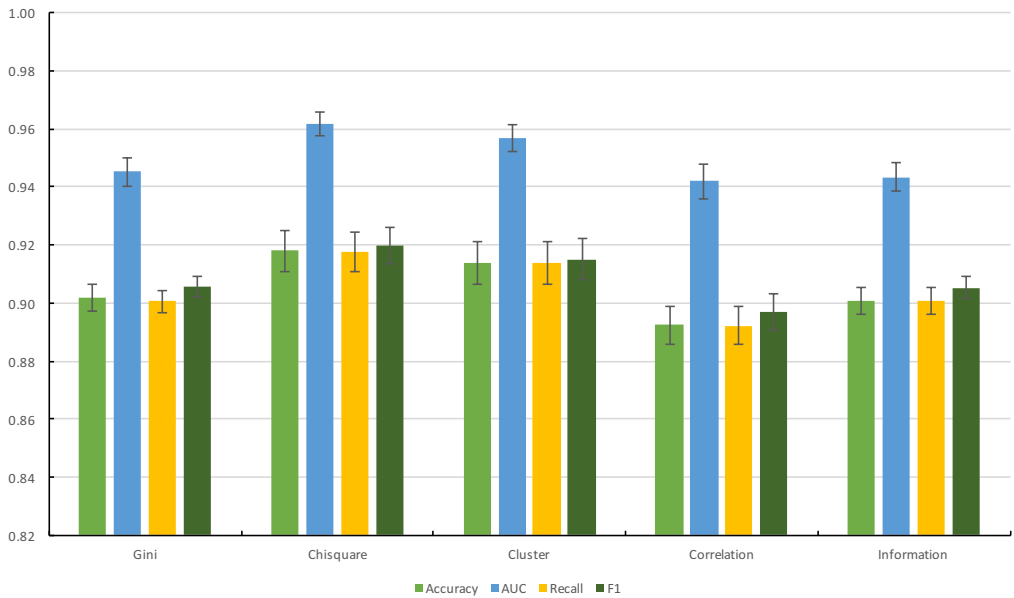

Figure 3. Bar plot of different evaluation criteria on XGB_TPE over five FS approaches

To further investigate and compare the effectiveness of different FS approaches, the Wilcoxon signed rank test was applied between each pair of the FS methods and the results are illustrated in Table 2. As described in Section 3.6, the Bonferroni correction significance level is set to $\alpha/10 = 0.1/10 = 0.01$ for the comparison and the $p$ value lower than 0.01 denotes the statistically significant. In general, different FS methods produce statistically significant difference since many $p$ values are smaller than 0.01. As expected, the performance difference between Chi-square and Cluster is not extremely large since the p values for the Wilcoxon signed rank test based on accuracy and recall are much larger than 0:01. Gini and Information produce statistically equal performance with respect to AUC, recall and F1. It is worth noting that Chi-square and Cluster outperforms the other three FS methods and the superiority is statistically significant. AUC obtained by Chi-square is significantly higher than Cluster. Furthermore, considering that Chi-square can result in slightly higher although not significantly higher accuracy, recall as well as F1 score than Cluster, Chi-square is selected as the optimal FS methods for XGB_TPE model.

### 4.3. Best FS method in XGB_RS

Similar as in XGB_TPE, we investigate the performance of XGB_RS after applying five different FS approaches. Later, the Wilcoxon signed rank test was applied between each pair of the FS methods using the four evaluation criteria. It is shown that the effect of FS approaches on XGB_RS performance is very similar as that on XGB_TPE. Therefore, the bar plot as well as the result of the Wilcoxon signed rank test is not listed in this paper. As a concise result, Chi-square is selected as the optimal FS method for the XGB_RS model.

### 4.4. Best FS method in LR

Fig. 4 demonstrated the LR performance over five FS approaches by using the four evaluation measures. By comparing Figs. 3 and 4, we found that the effect of FS approaches on evaluation measures is model-dependent. For example, by using Gini in XGB_TPE, an acceptable recall can be obtained. However, Gini results in the lowest recall in LR model. Correlation exhibits a promising AUC in LR while achieves the lowest AUC in XGB_TPE. Compared with the rest of the FS methods, although Cluster achieves the second highest F1 score in XGB_TPE, it results the lowest F1 score in LR. In XGB_TPE, AUC varies significantly across different FS methods while the change is not obvious in LR.

The Wilcoxon signed rank test was then applied between each pair of the five FS methods. For simplicity, the results of the Wilcoxon signed rank test for LR is not listed in this paper but the general conclusions obtained are shown as follows. Cluster demonstrates the best recall performance while Correlation has the worst result. Gini, Correlation, and Information do not seem to be promising FS methods compared to Chi-square and Cluster because of their relatively lower recall values. The effect on AUC caused by different FS methods is not obvious since except Cluster, there is no significant difference in accuracy between any pairs of the rest four FS approaches. Although Chi-square achieves the highest accuracy value, this method cannot result in equally high recall as Cluster. It is worth noting that Cluster has the worst performance by considering accuracy and F1 measures, although this method has the best performance recall and the second largest AUC. Although Cluster demonstrates lower accuracy than Chi-square, the difference is not statistically significance at the significant level of 0:05. Considering the importance of recall in this paper, Cluster is selected as the optimal FS method for LR model.

### 4.5. Final model selection

As discussed in Section 3, the final goal of this study aims at exploring the optimal model for business risk prediction. Therefore, the performance of LR, XGB_RS, and XGB_TPE are compared after selecting the best FS method for each model. Fig. 5 demonstrated the boxplot of the model performance based on their own best FS methods. The x-axis represents each of the three models, and the y-axis denotes accuracy, AUC, recall, and F1 score from the top left to the bottom right, respectively. It is found that XGB models (both XGB_RS and

XGB_TPE) outperform the traditional LR in all the four evaluation measures. XGB_TPE, which bases on Bayesian hyper-parameter optimization approach, achieves a higher accuracy, recall, and F1 score than XGB_RS that bases on a random trial-and-error process. The difference of AUC is not obvious between XGB_RS and XGB_TPE.

Table 2. Results of Wilcoxon signed rank test between four FS methods based on different criteria. $\alpha$ value is after Bonferroni correction.

| FS method | Criterion | p value | $\alpha = 0.01$ | Criterion | p value | $\alpha = 0.01$ |
|---|---|---|---|---|---|---|
| Gini vs. Chi-square | Accuracy | 0.0029 | Rejected | AUC | 0.0010 | Rejected |
| Gini vs. Cluster | | 0.0039 | Rejected | | 0.0010 | Rejected |
| Gini vs. Correlation | | 0.0010 | Rejected | | 0.0010 | Rejected |
| Gini vs. Information | | 0.0096 | Rejected | | 0.0527 | Not rejected |
| Chi-square vs. Correlation | | 0.0570 | Not rejected | | 0.0010 | Rejected |
| Chi-square vs. Information | | 0.0029 | Rejected | | 0.0010 | Rejected |
| Chi-square vs. Correlation | | 0.0029 | Rejected | | 0.0010 | Rejected |
| Chi-square vs. Information | | 0.0010 | Rejected | | 0.0010 | Rejected |
| Cluster vs. Information | | 0.0039 | Rejected | | 0.0010 | Rejected |
| Correlation vs. Information | | 0.0020 | Rejected | | 0.1162 | Not rejected |

| FS method | Criterion | p value | $\alpha = 0.01$ | Criterion | p value | $\alpha = 0.01$ |
|---|---|---|---|---|---|---|
| Gini vs. Chi-square | Recall | 0.0010 | Rejected | F1 score | 0.0010 | Rejected |
| Gini vs. Cluster | | 0.0020 | Rejected | | 0.0020 | Rejected |
| Gini vs. Correlation | | 0.0029 | Rejected | | 0.0010 | Rejected |
| Gini vs. Information | | 0.3848 | Not rejected | | 0.3050 | Not rejected |
| Chi-square vs. Correlation | | 0.0654 | Not rejected | | 0.0322 | Not rejected |
| Chi-square vs. Information | | 0.0010 | Rejected | | 0.0010 | Rejected |
| Chi-square vs. Correlation | | 0.0010 | Rejected | | 0.0010 | Rejected |
| Chi-square vs. Information | | 0.0010 | Rejected | | 0.0010 | Rejected |
| Cluster vs. Information | | 0.0010 | Rejected | | 0.0054 | Rejected |
| Correlation vs. Information | | 0.0010 | Rejected | | 0.1162 | Not rejected |

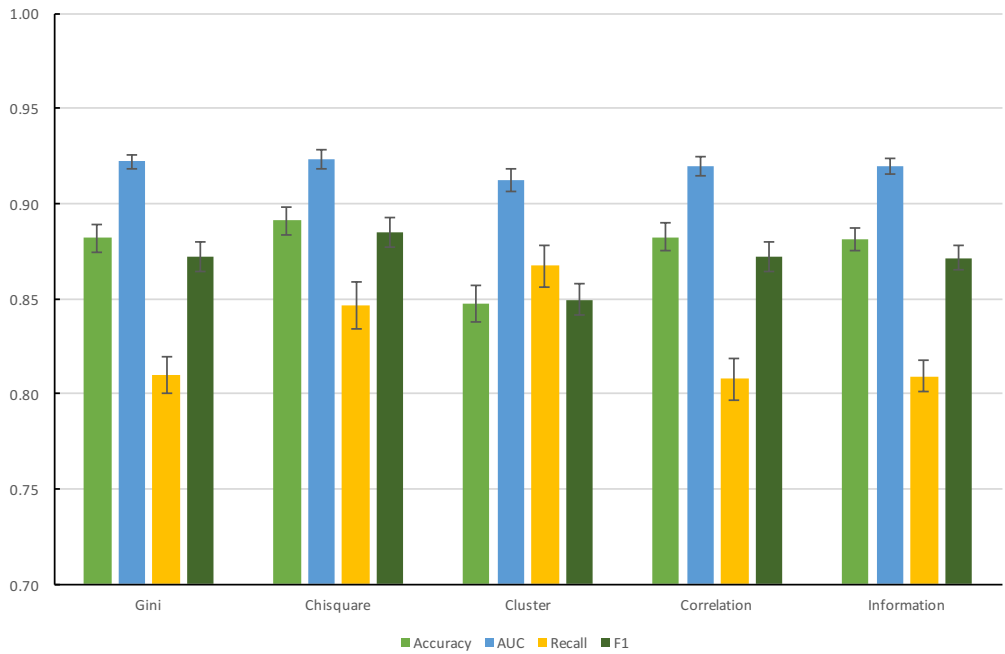

Figure 4. Bar plot of different evaluation criteria on LR over five FS approaches

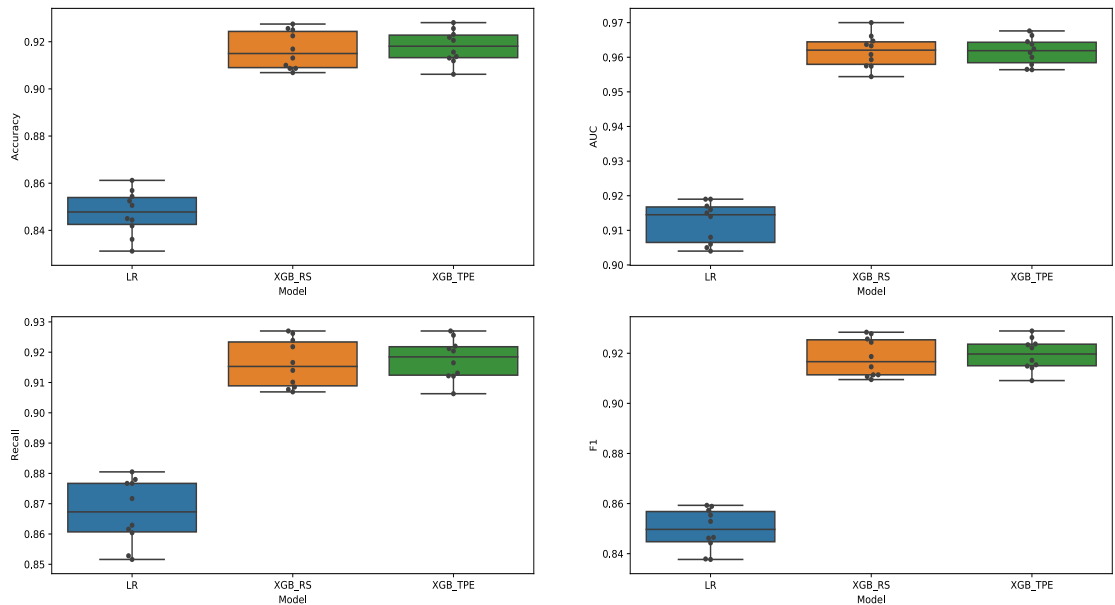

Figure 5. Box plot of different evaluation criteria across different models based on their own best FS approaches

To further examine the difference among LR, XGB_RS and XGB_TPE, the pairwise Wilcoxon signed rank test is then implemented. The results of the test are displayed in Table 3. As described in Section 3.6, the Bonferroni correction significance level is set to α/3 =

0.1/3 = 0.033 for the comparison and the p value lower than 0.033 denotes the statistically significant. As shown in Table 3, XGB methods (both XGB_RS and XGB_TPE) performs significantly better than LR. XGB_TPE is marginally better than XGB_RS since XGB_TPE significantly exceeds XGB_RS only in accuracy. Another finding is that, as depicted in Fig. 5, XGB_TPE shows a lower variability than XGB_RS with respect to accuracy, recall, and F1 score in our experiment. By contrast, XGB_RS depicts an even larger variability than LR and XGB_TPE if considering accuracy, recall, and F1 measures. Combining these aforementioned findings, we recommend XGB_TPE as the optimal model for business risk prediction in this study.

Table 3. Results of Wilcoxon signed rank test between each pair of LR, XGB_RS, XGB_TPE models. $\alpha$ value is after Bonferroni correction.

| FS method | Criterion | p value | $\alpha = 0.033$ | Criterion | p value | $\alpha = 0.033$ |
|---|---|---|---|---|---|---|
| LR vs. XGB_RS | Accuracy | 0.0010 | Rejected | AUC | 0.0010 | Rejected |
| LR vs. XGB_TPE | | 0.0010 | Rejected | | 0.0010 | Rejected |
| XGB_RS vs. XGB_TPE | | 0.0961 | Rejected | | 0.5000 | Not rejected |

| FS method | Criterion | p value | $\alpha = 0.033$ | Criterion | p value | $\alpha = 0.033$ |
|---|---|---|---|---|---|---|
| LR vs. XGB_RS | Recall | 0.0010 | Rejected | F1 score | 0.0010 | Rejected |
| LR vs. XGB_TPE | | 0.0010 | Rejected | | 0.0010 | Rejected |
| XGB_RS vs. XGB_TPE | | 0.1162 | Not rejected | | 0.1377 | Not rejected |

### 4.6. Rank of variable importance

After selecting the optimal candidate model, the importance of the variables is ranked to increase the model interpretability. Fig. 6 shows the top 15 most important features after training the XGBoost model with TPE hyper-parameter optimization. Higher F score imply the more importance of the corresponding features. Therefore, the feature pctNFChgAccAcc24mon (i.e., percent of non-financial charge-o accounts to total accounts reported in last 24 months) is the most important variable in the risk prediction and should be highlighted in the collection of the credit data. By contrast, pctNFPDAmt24mon (i.e., percent of non-financial past due amount to total balance reported in last 24 months) shows a lower necessity in the model.

### 5. CONCLUSION AND DISCUSSION

In this study, we introduce XGBoost, one of the state-of-the-art machine learning techniques, into the business risk modeling domain, aiming to explore a more accurate business risk model compared to the standard LR. Moreover, the FS methods and hyper-parameter optimization are examined simultaneously in the modeling procedure. The dataset used in our study contains the commercial information from over 10 million of the de-identified companies in the U.S. from 2006 to 2014. Our experiments are repeated 10 times of the 10-fold cross validation. Five FS methods including Gini, Chi-square, Cluster, Correlation, and Information are employed to remove redundant variables. Two hyper-parameter tuning methods including RS and TPE are used in XGBoost. Finally, the effects of FS and hyper-parameter tuning methods on the model performance are comprehensively investigated by the pairwise Wilcoxon signed rank test.

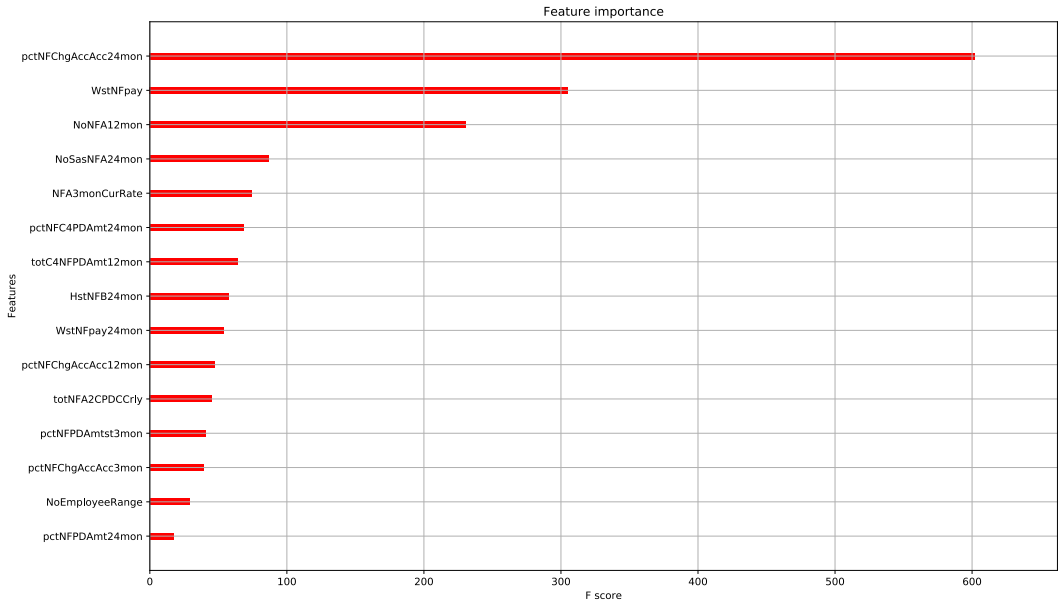

Figure 6. Top 15 most important features based on XGB_TPE model

Our analysis shows that the effect of FS methods on the model performance dependents on the model type. In LR, Gini FS method can result in the lowest recall while it exhibits an acceptable recall in XGB (both XGB_RS and XGB_TPE). Different FS methods result in significant changes in AUC for XGB models but do not have obvious effect in LR. The Cluster FS method is shown to be the optimal FS methods for LR while Chi-square outperforms other FS methods in both XGB_RS and XGB_TPE. The comparisons with traditional LR show the significant superiority of the XGBoost methods (both XGB_RS and XGB_TPE) in terms of accuracy, AUC, recall and F1 score. Bayesian TPE hyper-parameter optimization method is significantly better than RS hyper-parameter tuning, since XGB_TPE achieves significantly higher accuracy than XGB_RS. Furthermore, XGB_TPE outperforms XGB_RS in terms of AUC, re-call, and F1 score, although the improvements are not statistically significant. It is also worth noting that XGB_TPE shows a lower variability than XGB_RS by considering accuracy, recall, and F1 score. As the final result, we conclude that XGB_TPE is marginally better than XGB_RS while significantly better than LR. Therefore, XGB_TPE is selected as the optimal model for business risk modeling in our study. The ranking of the variable importance shows that pctNFChgAccAcc24mon is the most important variable in the risk predictions while the weight of pctNFPDAmt24mon is not obvious in the final model. The result demonstrated in Fig. 6 can provide guidance to financial institutes in the collection of credit data.

Besides the above-mentioned promising results achieved by XGBoost on risk modeling using the medium sized data in this study, XGBoost has been demonstrated to be powerful in handling large scale data using very limited computing resources [20]. According to the experimental results in [20], XGBoost achieves scalable learning through parallel and distributed computing, out-of-core computation, and cache-aware learning. When the real-world data used in the risk modeling domain is large, the out-of-core computation in XGBoost can utilize the disk space if the data is too large to fit into the main memory. Therefore, XGBoost provides the insights for the data scientist on how to efficiently manage and load large scale database using minimal amount of computing resources.

In the future business risk modeling studies, the results might not be consistent because of using different dataset. However, the workflow proposed in our study may serve as a reference for future studies in building XGBoost models and ranking variable importance in the credit domain. This study can also provide a guidance for comprehensively exploring the effect of FS algorithms as well as hyper-parameter optimization on the model performance.

**Authors**

Yan Wang is a Ph.D. candidate in Analytics and Data Science at Kennesaw State University. Her research interest contains algorithms and applications of data mining and machine learning techniques in financial areas. She has been a summer Data Scientist intern at Ernst & Young and focuses on the fraud detections using machine learning techniques. Her current research is about exploring new algorithms/models that integrates new machine learning tools into traditional statistical methods, which aims at helping financial institutions make better strategies. Yan received her M.S. in Statistics from university of Georgia. 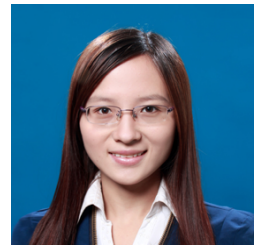

Dr. Xuelei Sherry Ni is currently a Professor of Statistics and Interim Chair of Department of Statistics and Analytical Sciences at Kennesaw State University, where she has been teaching since 2006. She served as the program director for the Master of Science in Applied Statistics program from 2014 to 2018, when she focused on providing students an applied leaning experience using real-world problems. Her articles have appeared in the Annals of Statistics, the Journal of Statistical Planning and Inference and Statistica Sinica, among others. She is the also the author of several book chapters on modeling and forecasting. Dr. Ni received her M.S. and Ph.D. in Applied Statistics from Georgia Institute of Technology. 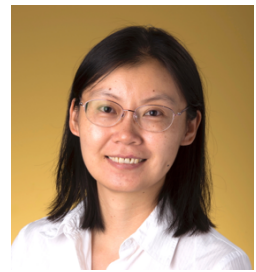